# Towards Enriched Controllability for Educational Question Generation

Bernardo Leite[1,2][0000−0002−9054−9501] and
Henrique Lopes Cardoso[1,2][0000−0003−1252−7515]

[1] Faculty of Engineering of the University of Porto (FEUP), Porto, Portugal
[2] Artificial Intelligence and Computer Science Laboratory (LIACC), Porto, Portugal
{bernardo.leite, hlc}@fe.up.pt

**Abstract.** *Question Generation* (QG) is a task within Natural Language Processing (NLP) that involves automatically generating questions given an input, typically composed of a text and a target answer. Recent work on QG aims to control the type of generated questions so that they meet educational needs. A remarkable example of *controllability* in educational QG is the generation of questions underlying certain *narrative elements*, e.g., causal relationship, outcome resolution, or prediction. This study aims to enrich controllability in QG by introducing a new guidance attribute: *question explicitness*. We propose to control the generation of explicit and implicit (*wh*)-questions from children-friendly stories. We show preliminary evidence of controlling QG via question explicitness alone and simultaneously with another target attribute: the question's narrative element. The code is publicly available at github.com/bernardoleite/question-generation-control.

## 1 Introduction

In the educational context, Question Generation (QG) can potentially automate and assist the teacher in what can be a time-consuming and effortful task. QG may also be helpful for the learner's formative assessment via self-study and engagement with computer-generated practice questions. However, automatic QG tools are not widely used in classrooms [2,8], namely because generated questions are generally limited in types and difficulty levels [2]. As pointed by Wang *et al.* [8], there is a strong desire for user control, where humans provide input to QG systems and can decide when to use their output. Inspired by this need, this study proposes a QG framework for controlling the generation of explicit and implicit questions, using question explicitness as a guidance attribute during the generation process. Generally, explicit questions center on a particular story fact, whereas implicit questions rely on summarizing[3] and drawing inferences from implicit information in the text. As stated by Xu *et al.* [9], explicit and implicit questions are formally defined as follows:

- **Explicit** *questions ask for answers that can be directly found in the stories. In other words, the source of answer are spans of text.*
- **Implicit** *questions ask for answers that cannot be directly found in the text. Answering the questions requires either reformulating language or making inferences. In other words, the answer source is "free-form", meaning that the answers can be any free-text, and there is no limit to where the answer comes from.*

---

[3] Summarization skills have been used to assess and improve students' reading comprehension ability [8].

Noteworthy, prior research [6,11,9] suggests that a combination of explicit and implicit questions contributes to a more balanced difficulty in the assessments. To achieve our goal, we use a recent dataset called FairytaleQA [9], which contains question-answering (QA) pairs derived from children-friendly stories. Each question is categorized as "explicit" or "implicit" by expert annotators.

Some previous work has addressed controllability in educational QG. For instance, Ghanem *et al.* [1] control the reading comprehension skills required by the question, e.g., figurative language and summarization. Similarly, Zhao *et al.* [10] control the narrative elements underlying the generated questions, such as causal relationship, outcome resolution, or prediction. They use the same dataset as this study, FairytaleQA, where each question, beyond explicitness, is also categorized according to the referred narrative elements.

## 2   Generating Explicit and Implicit Questions

In this study, we fine-tune the T5 pre-trained model [5] with the controllable mechanism for generating explicit and implicit questions. T5 is a text-to-text generation model which has achieved state-of-the-art results on multiple natural language generation benchmarks, including QA and summarization. We train the model to generate both questions and answers for a particular story text. To control the explicitness of the generated questions, we prepend a special token <EX> followed by "explicit" or "implicit" attribute at the beginning of the input, before the story text. This attribute guides the system to generate a question of the desired type. Other special tokens (<SECTION>, <QUESTION> and <ANSWER>) are used to delimit the input and output information of the model. This technique is based on a recent study [10] with the purpose of controlling QG conditioned on another target attribute: the question's narrative elements. We also investigate controlling simultaneously the question's explicitness along with that target attribute. To that end, beyond <EX>, we prepend <NAR> followed by the narrative attribute name.

## 3   Experimental Setup

**Data**: We use FairytaleQA [9], in which educational experts have manually created 10,580 QA pairs from 278 children-friendly stories. Each question is annotated with an explicitness label, which can be "explicit" or "implicit". Also, each question is labeled with one of the following narrative elements[4]: "character", "setting", "action", "feeling", "causal relationship", "outcome resolution", or "prediction". Statistically, each story has ≈15 sections and each section (composed of multiple sentences) has ≈3 questions. Explicit questions represent ≈75% of all questions. We use the original train/val/test splits composed of 8,548/1,025/1,007 QA pairs.

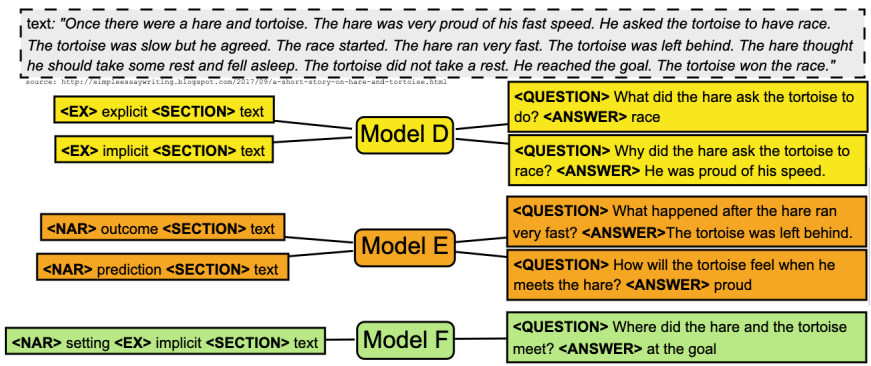

**Fig. 1.** An illustrative example of question-answer pairs generated by different models.

**Models**: From the original dataset, we have trained different models[5]: (A) question-section:answer; (B) answer-section:question; (C) section:question-answer; (D) ex-section:question-answer; (E) nar-section:question-answer; and (F) nar-ex-section:question-answer. Models A and B will serve as a baseline comparison with the QA and QG models from the FairytaleQA paper. Model C only contains the section text as input, so its purpose is to serve as a baseline to compare with models D-F, which include control attributes. Model D includes the question's explicitness attribute in the input. Model E includes the narrative attribute in the input. Model F has both control attributes included. Figure 1 shows an illustrative example of the models with controllability prompts.

***controlled test* set**: For assessing the effectiveness of controllability along models D-F, we have prepared a reorganized version from the original *test* set which we call *controlled test*: each example includes a section and all ground-truth QA pairs regarding that section, being that these QA pairs belong to one explicitness type (explicit or implicit) and narrative element. Also, for comparability between models C and D-F, each section only appears once.

**Implementation Details**: We use the *t5-base*[6] model version. We have set 512 and 128 for the maximum token input and output, respectively. We train the models with a maximum of 10 epochs, early stopping with a patience of 2, and a batch size of 32. For inference, we use beam search with a beam width of 5.

---

[4] Detailed information of each aspect is described in the FairytaleQA paper [9].
[5] A colon separates the input and output information used by the models.
[6] https://huggingface.co/t5-base

## 4   Results

**Baselines**: FAIRYTALEQA authors have reported $n$-gram similarity $\text{ROUGE}_L$-F1 [3] values of 0.536 (QA) and 0.527 (QG) on the *test* set. Using our baseline models (A and B) we correspondingly obtained 0.559 (QA) and 0.529 (QG). This shows that our baseline models are quantitatively aligned with previously obtained results.

**QA results by question explicitness**: More on baseline model A for QA, our $\text{ROUGE}_L$-F1 QA results for explicit and implicit questions are 0.681 and 0.194, respectively. This notable difference is also observe by Xu *et al.* [9]. According to the authors, this situation is expected since the answers to explicit questions can be directly found in the text. In contrast, implicit questions call for in-depth inference and summarization. We use this rationale to evaluate the controllability of the question's explicitness. We hypothesize that the QA model obtained in setup A will perform significantly better on explicit than implicit questions generated from models D and F.

**Controllability**: We look for evidence of the question's controllability by employing both QA and QG tasks. For QA, we use the $\text{ROUGE}_L$-F1 metric and EXACT MATCH, which is a strict all-or-nothing score between two strings. For QG, we use $n$-gram similarity $\text{ROUGE}_L$-F1 and BLEU-4 [4]. Also, we use BLEURT [7], which is a more recent text generation performance metric.

Table 1 refers to the QA results, which have been obtained as follows. We use the QA model (obtained in setup A) for answering the generated questions from models D and F. Then, the answers obtained from the QA model are compared against the answers generated from models D and F, yielding the reported results. For both evaluation metrics, the QA model performs significantly better on explicit than implicit generated questions (confirming our hypothesis). Thus, we conclude that these scores indicate compelling evidence that it is possible to control the question's explicitness using the proposed controllable mechanism.

Table 2 presents the obtained QG results. Here the traditional evaluation procedure in QG is employed, which is to directly compare the generated questions with the ground-truth[7]. We find no significant differences in the QG scores obtained by model D compared to C, which can be explained as follows: controlling the question's explicitness has more influence on the type of answer required to respond to the generated question than on the syntax of that generated question. Therefore, we consider the non-significant differences between models C and D in the QG results to be expected. In contrast, a significant improvement is observed in models E and F (which receive narrative controllability prompts) compared to model C. This can be explained as follows: controlling the question's narrative elements strongly influences the syntax of the generated questions. For instance, we empirically observe that when requesting the model to generate questions about the "causal relationship" element, it generates (in many cases) questions starting with "Why did...?". As for "outcome resolution", the model generates "What happened...?" questions. As for "prediction", the model generates "How will...?" questions.

---

[7] Note that the drop in QG $\text{ROUGE}_L$-F1 values relative to baseline model B is expected, since in these models the answer is not included in the input. The generated questions may thus focus on target answers that are not part of the gold standard.

Finally, it should be noted that model F (which receives both explicitness and narrative controllability prompts) is shown to be effective for controlling *simultaneously* question's explicitness and question's narrative elements.

**Table 1.** QA results (0-1) for assessing the question's controllability (*controlled test*).

|  | **ROUGE$_L$-F1** | | | **EXACT MATCH** | | |
| :---: | :---: | :---: | :---: | :---: | :---: | :---: |
| **Models** | Overall | Explicit | Implicit | Overall | Explicit | Implicit |
| ex-section:question-answer (D) | 0.656 | 0.741 | 0.431 | 0.434 | 0.483 | 0.306 |
| nar-ex-section:question-answer (F) | 0.671 | 0.730 | 0.514 | 0.449 | 0.489 | 0.343 |

**Table 2.** QG results (0-1) for assessing the question's controllability (*controlled test*).

| Models | ROUGE$_L$-F1 | BLEU-4 | BLEURT |
| :---: | :---: | :---: | :---: |
| section:question-answer (C) | 0.305 | 0.099 | 0.370 |
| ex-section:question-answer (D) | 0.303 | 0.104 | 0.369 |
| nar-section:question-answer (E) | 0.432 | 0.189 | 0.432 |
| nar-ex-section:question-answer (F) | 0.432 | 0.195 | 0.424 |

## 5 Conclusion

In this study, we work towards enriched controllability for educational QG. Through automatic evaluation, the results show preliminary evidence that it is possible to (1) control the question's explicitness and (2) *simultaneously* control both the question's explicitness and question's narrative elements. We argue that the next developments in educational QG should involve enriching (even more) the controllability process with multiple guidance and educationally relevant attributes. Looking for additional effective control mechanisms is also an interesting route. For future work, we intend to perform a large-scale human evaluation focusing on QG controllability in an actual educational environment.

**Acknowledgments** This work was financially supported by Base Funding - UIDB/00027/2020 of the Artificial Intelligence and Computer Science Laboratory – LIACC - funded by national funds through the FCT/MCTES (PIDDAC). Bernardo Leite is supported by a PhD studentship (with reference 2021.05432.BD), funded by Fundação para a Ciência e a Tecnologia (FCT).

## References

1. Ghanem, B., Lutz Coleman, L., Rivard Dexter, J., von der Ohe, S., Fyshe, A.: Question generation for reading comprehension assessment by modeling how and what to ask. In: Findings of the Association for Computational Linguistics: ACL 2022. pp. 2131–2146. Association for Computational Linguistics, Dublin, Ireland (May 2022). https://doi.org/10.18653/v1/2022.findings-acl.168
2. Kurdi, G., Leo, J., Parsia, B., Sattler, U., Al-Emari, S.: A systematic review of automatic question generation for educational purposes. International Journal of Artificial Intelligence in Education **30**(1), 121–204 (2020). https://doi.org/10.1007/s40593-019-00186-y
3. Lin, C.Y.: ROUGE: A Package for Automatic Evaluation of Summaries. In: Text Summarization Branches Out. pp. 74–81. ACL, Barcelona, Spain (Jul 2004), https://www.aclweb.org/anthology/W04-1013
4. Papineni, K., Roukos, S., Ward, T., Zhu, W.J.: Bleu: a Method for Automatic Evaluation of Machine Translation. In: Proceedings of the 40th Annual Meeting of the Association for Computational Linguistics. pp. 311–318. ACL, Philadelphia, Pennsylvania, USA (Jul 2002). https://doi.org/10.3115/1073083.1073135
5. Raffel, C., Shazeer, N., Roberts, A., Lee, K., Narang, S., Matena, M., Zhou, Y., Li, W., Liu, P.J.: Exploring the limits of transfer learning with a unified text-to-text transformer. Journal of Machine Learning Research **21**(140), 1–67 (2020), http://jmlr.org/papers/v21/20-074.html
6. Raphael, T.E.: Teaching question answer relationships, revisited. The reading teacher **39**(6), 516–522 (1986)
7. Sellam, T., Das, D., Parikh, A.: BLEURT: Learning robust metrics for text generation. In: Proceedings of the 58th Annual Meeting of the Association for Computational Linguistics. pp. 7881–7892. Association for Computational Linguistics, Online (Jul 2020). https://doi.org/10.18653/v1/2020.acl-main.704
8. Wang, X., Fan, S., Houghton, J., Wang, L.: Towards process-oriented, modular, and versatile question generation that meets educational needs. In: Proceedings of the 2022 Conference of the North American Chapter of the Association for Computational Linguistics: Human Language Technologies. pp. 291–302. Association for Computational Linguistics, Seattle, United States (Jul 2022). https://doi.org/10.18653/v1/2022.naacl-main.22
9. Xu, Y., Wang, D., Yu, M., Ritchie, D., Yao, B., Wu, T., Zhang, Z., Li, T., Bradford, N., Sun, B., Hoang, T., Sang, Y., Hou, Y., Ma, X., Yang, D., Peng, N., Yu, Z., Warschauer, M.: Fantastic questions and where to find them: FairytaleQA – an authentic dataset for narrative comprehension. In: Proceedings of the 60th Annual Meeting of the Association for Computational Linguistics (Volume 1: Long Papers). pp. 447–460. Association for Computational Linguistics, Dublin, Ireland (May 2022). https://doi.org/10.18653/v1/2022.acl-long.34
10. Zhao, Z., Hou, Y., Wang, D., Yu, M., Liu, C., Ma, X.: Educational question generation of children storybooks via question type distribution learning and event-centric summarization. In: Proceedings of the 60th Annual Meeting of the Association for Computational Linguistics (Volume 1: Long Papers). pp. 5073–5085. Association for Computational Linguistics, Dublin, Ireland (May 2022). https://doi.org/10.18653/v1/2022.acl-long.348
11. Zucker, T.A., Justice, L.M., Piasta, S.B., Kaderavek, J.N.: Preschool teachers' literal and inferential questions and children's responses during whole-class shared reading. Early Childhood Research Quarterly **25**(1), 65–83 (2010). https://doi.org/https://doi.org/10.1016/j.ecresq.2009.07.001